\documentclass[10pt,twocolumn,letterpaper]{article}

\usepackage{cvpr}              

\usepackage[accsupp]{axessibility}
\usepackage{multirow}
\usepackage{makecell}
\usepackage{bigstrut}
\usepackage[ruled,linesnumbered]{algorithm2e}

\usepackage[dvipsnames]{xcolor}

\definecolor{cvprblue}{rgb}{0.21,0.49,0.74}
\usepackage[pagebackref,breaklinks,colorlinks,citecolor=cvprblue]{hyperref}

\def\paperID{6482} 
\def\confName{CVPR}
\def\confYear{2024}

\title{Latent Modulated Function for Computational Optimal\\Continuous Image Representation}

\author{Zongyao He\textsuperscript{1} \hspace{0.5cm} Zhi Jin\textsuperscript{1,}\textsuperscript{2,}\thanks{Corresponding author}\\
\textsuperscript{1}School of Intelligent Systems Engineering, Shenzhen Campus of Sun Yat-sen University\\ \textsuperscript{2}Guangdong Provincial Key Laboratory of Fire Science and Technology\\
{\tt\small {hezy28@mail2.sysu.edu.cn, jinzh26@mail.sysu.edu.cn}}
}

\begin{document}

\maketitle
\begin{abstract}
The recent work Local Implicit Image Function (LIIF) and subsequent Implicit Neural Representation (INR) based works have achieved remarkable success in Arbitrary-Scale Super-Resolution (ASSR) by using MLP to decode Low-Resolution (LR) features. However, these continuous image representations typically implement decoding in High-Resolution (HR) High-Dimensional (HD) space, leading to a quadratic increase in computational cost and seriously hindering the practical applications of ASSR.
To tackle this problem, we propose a novel Latent Modulated Function (LMF), which decouples the HR-HD decoding process into shared latent decoding in LR-HD space and independent rendering in HR Low-Dimensional (LD) space, thereby realizing the first computational optimal paradigm of continuous image representation.
Specifically, LMF utilizes an HD MLP in latent space to generate latent modulations of each LR feature vector. This enables a modulated LD MLP in render space to quickly adapt to any input feature vector and perform rendering at arbitrary resolution.
Furthermore, we leverage the positive correlation between modulation intensity and input image complexity to design a Controllable Multi-Scale Rendering (CMSR) algorithm, offering the flexibility to adjust the decoding efficiency based on the rendering precision. Extensive experiments demonstrate that converting existing INR-based ASSR methods to LMF can reduce the computational cost by up to $99.9\%$, accelerate inference by up to $57 \times$, and save up to $76\%$ of parameters, while maintaining competitive performance. The code is available at \href{https://github.com/HeZongyao/LMF}{https://github.com/HeZongyao/LMF}.
\end{abstract}  

\begin{figure}[t]
  \centering
  \includegraphics[width=0.98\linewidth]{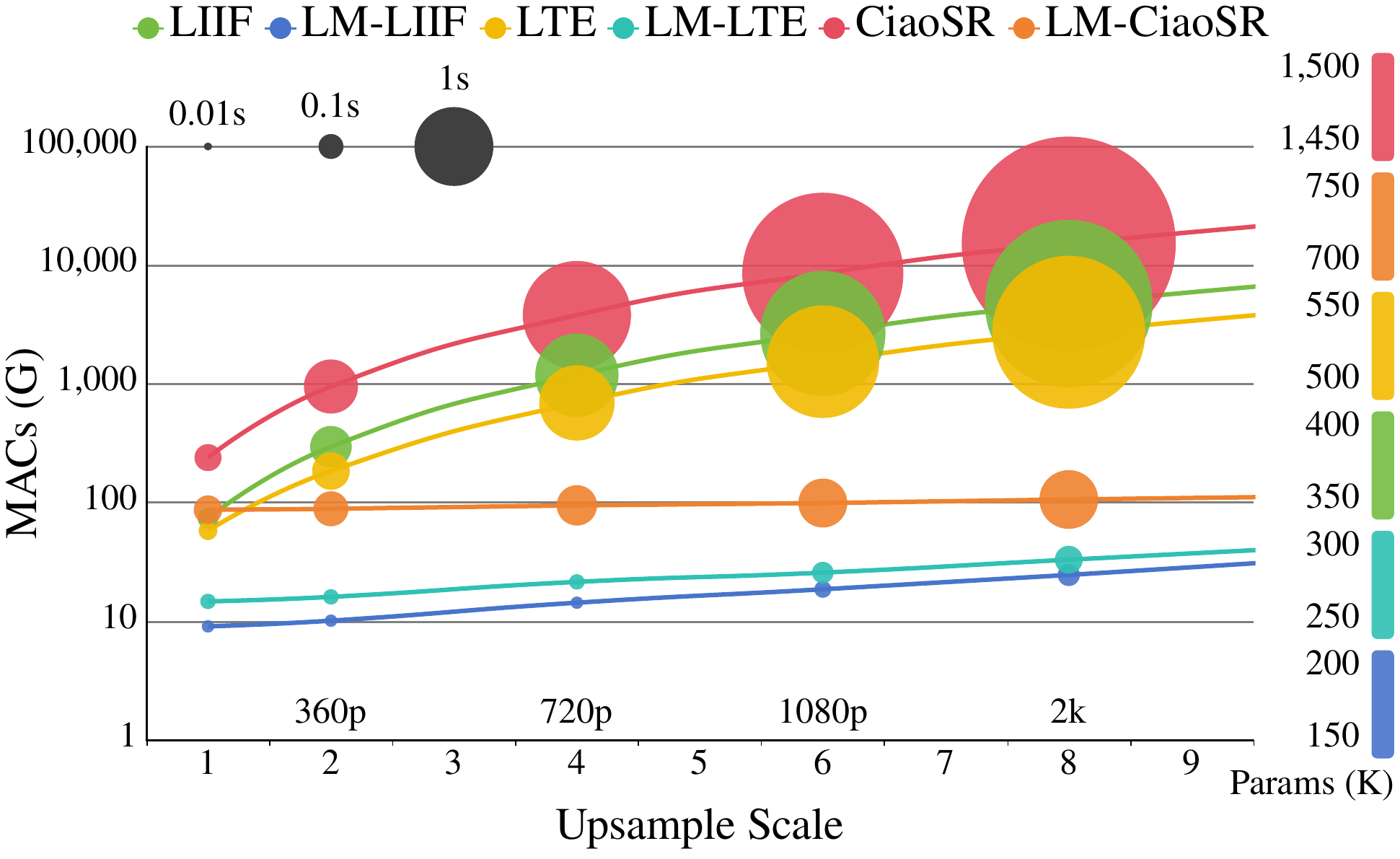}
  \caption{Efficiency comparisons ($320\times180$ input) for ASSR. LMF-based ASSR methods significantly reduce computational cost (MACs), runtime (circle sizes), and parameters (colors).}
  \label{fig:teaser-figure}
  \vspace{-0.45cm}
\end{figure}

\begin{figure*}[ht]
  \centering
  \includegraphics[width=0.98\textwidth]{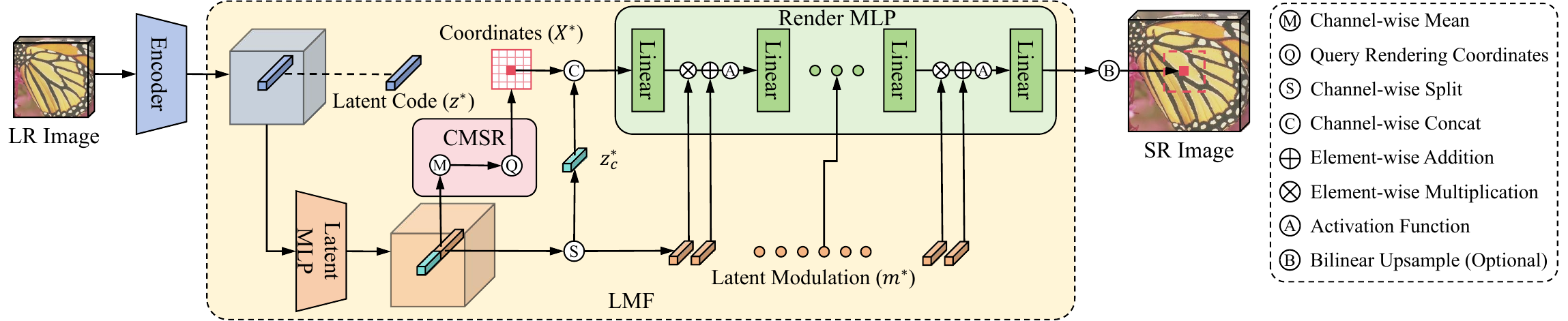}
  \caption{The framework of our LMF-based continuous image representation. Given a latent code generated by the encoder, an HD latent MLP generates the latent modulation to adjust the features of an LD render MLP, thereby achieving efficient arbitrary-resolution rendering.}
  \label{fig:lmif-framework}
  \vspace{-0.45cm}
\end{figure*}

\section{Introduction}
Image data has long been stored and presented as discrete 2D arrays with fixed resolutions, while the visual world we perceive is continuous and precise. As a result, there is a growing demand for arbitrary scale image resizing in various real-world scenarios. For instance, users often require continuous resizing of images during capture, viewing, and editing. Recent developments in Single Image Super-Resolution (SISR) \cite{dong2014learning,lai2017deep,ledig2017photo,lim2017enhanced,tai2017memnet,zhang2018image,zhang2018residual,li2019feedback,zhang2020multi,he2021srdrl,chen2021pre,liang2021swinir,chen2023hat} have showcased significant advancements in reconstructing High-Resolution (HR) images from Low-Resolution (LR) counterparts. However, traditional interpolation methods are still the preferred choice for handling real-world image resizing. This is because SISR methods typically learn upsampling functions for a particular scale factor, requiring training different models for different scales. In contrast, Arbitrary-Scale Super-Resolution (ASSR) utilizes a single model to handle arbitrary scales, thus presenting significant potential in practical image-resizing applications.

Building on the success of Implicit Neural Representation (INR), Local Implicit Image Function (LIIF) \cite{chen2021learning} is a recent ASSR approach that learns to continuously represent 2D images through an implicit function. LIIF encodes an image as latent codes and employs a decoding MLP to predict pixel values based on coordinates and their neighboring latent codes. 
The use of continuous coordinates enables LIIF and subsequent INR-based ASSR works \cite{wang2021learning, yang2021implicit, xu2021ultrasr, lee2022local, cao2023ciaosr, song2023ope, chen2023cascaded, wang2023deep, gao2023implicit, yao2023local, he2023dynamic} to represent images at any resolution, achieving robust performance in up to $\times 30$ SR task. However, existing INR-based ASSR methods adhere to an intuitive ASSR paradigm that uses a complex decoding function in HR High-Dimensional (HD) space to independently predict the pixel value of each HR coordinate. As a result, their computational cost and runtime increase quadratically with the scale factor, contradicting the goal of real-world image-resizing applications (refer to Figure \ref{fig:teaser-figure}).
Upon closer examination, the significant computational redundancy of HR-HD decoding can be divided into three aspects. Firstly, there is inherent similarity within the HR region corresponding to an LR latent code, suggesting the need for merging shared computations within this region before independently predicting each HR pixel value.
Secondly, this paradigm ignores the varying complexity of different LR latent codes and applies the same computational process to predict all the HR pixel values. Lastly, this paradigm couples the HD MLP parameters and HR computations, while such a large number of parameters are not always needed for each prediction.

To tackle this problem, we propose a novel Latent Modulated Function (LMF) to realize a computational optimal paradigm of continuous image representation. By leveraging latent modulation, LMF successfully disentangles the decoding process from HR-HD space into latent (LR-HD) and render (HR-LD) spaces. As shown in Figure \ref{fig:lmif-framework}, LMF first utilizes a latent MLP to generate the latent modulation of a given latent code. When querying the coordinates near the latent code, the latent modulation is applied to adjust each hidden linear layer in the render MLP. This modulation enables LMF to minimize the dimension of the render MLP without sacrificing performance.
Furthermore, we observe a positive correlation between the latent modulation intensity and the input image complexity. Building on this insight, we propose a Controllable Multi-Scale Rendering (CMSR) algorithm. CMSR not only allows the computational cost of rendering to be proportional to the input image complexity rather than the output resolution, but also provides the flexibility to balance precision and efficiency during testing.
Computational optimal and fast ASSR holds great promise for future research and downstream applications. Experiments demonstrate that transforming existing INR-based ASSR methods into LMF reduces computational costs by $90.4\%$ to $99.9\%$, accelerates inference by $2.4 \times$ to $56.9 \times$, and saves $45.1\%$ to $76.0\%$ of parameters, while maintaining competitive PSNR performance. 

Our main contributions are summarized as follows:
\begin{itemize}
\item We propose a Latent Modulated Function (LMF) that realizes computational optimal continuous image representation, which decouples the HR-HD decoding into decoding in latent (LR-HD) and render (HR-LD) spaces.

\item Based on the correlation between modulation intensity and input image complexity, we propose a Controllable Multi-Scale Rendering (CMSR) algorithm to effortlessly balance the rendering efficiency and precision at test time.

\item Extensive experiments demonstrate that seamlessly converting existing INR-based ASSR methods to LMF can achieve substantial reductions in computational cost, runtime, and parameters, while preserving performance.
\end{itemize} 
\section{Related Work}
\subsection{Implicit Neural Representation}
INR has demonstrated remarkable success for continuous signal representation across various tasks and modalities, including signal reconstruction \cite{sitzmann2020implicit, mehta2021modulated, saragadam2022miner, chen2022transformers}, image and video compression \cite{strumpler2022implicit, dupont2022coin++, chen2021nerv}, and image generation \cite{skorokhodov2021adversarial, yu2022generating}. The highlight of INR research is NeRF \cite{mildenhall2021nerf} and subsequent works \cite{barron2021mip,yu2021pixelnerf,barron2022mip,reiser2021kilonerf}, which learn MLPs to infer the RGB and density of any given coordinates and view directions in a 3D scene. However, a major challenge faced by INR is the extensive training process required to fit a signal from scratch. To address this, MINER \cite{saragadam2022miner} and KiloNeRF \cite{reiser2021kilonerf} propose the use of multiple small MLPs to independently learn different segments of a signal. Another solution is hyper-base network design \cite{skorokhodov2021adversarial,mehta2021modulated,dupont2022coin++,chen2022transformers}, where a pre-trained hyper network generates INR for any input signal during testing. To represent complex signals, hyper networks often generate modulations for base networks.

\subsection{Single-Image Super-Resolution}
SISR as a classic low-level vision task has been studied for decades. In recent years, this field witnessed a variety of CNN-based networks, such as SRCNN \cite{dong2014learning}, LapSRN \cite{lai2017deep}, SRResNet \cite{ledig2017photo}, EDSR \cite{lim2017enhanced}, MemNet \cite{tai2017memnet}, RCAN \cite{zhang2018image}, RDN \cite{zhang2018residual}, and SRFBN \cite{li2019feedback}. Lately, image processing transformers, such as IPT \cite{chen2021pre}, SwinIR \cite{liang2021swinir}, and HAT \cite{chen2023hat}, have been proposed, which surpass CNN-based methods in SR performance using a large dataset. Although SISR methods have achieved remarkable performance, most of them are designed for a specific scale.

ASSR methods are designed to utilize a single network for all scales, showcasing significant advancements in practicality and convenience over previous works. Hu \etal proposed MetaSR \cite{hu2019meta} to generate an upsampling layer for ASSR using a meta-network. However, MetaSR exhibits poor generalization for large-scale SR. Building upon the success of INR, a novel continuous image representation was proposed in LIIF \cite{chen2021learning}, where each pixel value is inferred from an MLP using its coordinate and neighboring 2D features. Subsequently, a variety of INR-based ASSR methods have been proposed \cite{wang2021learning, yang2021implicit, xu2021ultrasr, lee2022local, cao2023ciaosr, song2023ope, chen2023cascaded, wang2023deep, gao2023implicit, yao2023local, he2023dynamic}. Notably, Lee \etal proposed LTE \cite{lee2022local}, which introduces a dominant-frequency estimator to continuously capture fine details. Cao \etal introduced CiaoSR \cite{cao2023ciaosr}, a continuous implicit attention-in-attention network that learns ensemble weights for nearby local features, achieving State-Of-The-Art (SOTA) performance. Despite the robust performance of INR-based ASSR methods for scales up to $\times30$, they adhere to the paradigm of decoding in HR-HD space, leading to prohibitively high computational cost and runtime.

\section{Methods}
\subsection{Preliminary and Motivation}
In continuous image representations, an image $I_{in}$ is represented as 2D feature maps $F$ in latent space $\mathbb{R}^{H_{in} \times W_{in} \times D_{F}}$. A latent code $z^{*} \in \mathbb{R}^{D_{F}}$ represents a feature vector derived from the feature maps, which is used to decode its nearest coordinates in the continuous spatial domain $X \subset \mathbb{R}^{2}$.
Since the encoder can be any feature extractor from existing SISR methods, the essence of continuous image representations is the decoding function, which arbitrarily upsamples LR-HD features to HR-LD pixel values. A simple and intuitive paradigm is to learn a decoding function in HR-HD space that independently decodes the pixel value of each HR coordinate. This vanilla decoding function $f_{\theta}$ parameterized as an HD MLP $\theta$ can be formulated as:
\begin{equation}
  I(x^{*}_{i, j}) = f_{\theta}(z^{*}, x^{*}_{i, j}),
  \label{con:liif}
\end{equation}
where $x^{*}_{i, j} \in X^{*}$ refers to any nearest coordinate of $z^{*}$, and $I(x^{*}_{i, j})$ represents the predicted pixel value at $x^{*}_{i, j}$.

However, as mentioned earlier in the Introduction section, this vanilla paradigm suffers from three aspects of computational redundancy.
A computational optimal decoding function should allow computations proportional to the input image complexity rather than the output resolution. With this in mind, we propose to decouple the HR-HD decoding process into shared decoding in latent space and independent rendering in render space, along with a rendering algorithm adaptive to the input image content.

\subsection{Latent Modulated Function}
To eliminate the computational redundancy in HR-HD decoding, we first propose a two-stage decoding process comprising a latent MLP $\theta_{l}$ and a render MLP $\theta_{r}$. The latent MLP handles the shared decoding process within the HR region of a latent code $z^{*}$, while the render MLP independently predicts the pixel value of each HR coordinate. The LR-HD output $f_{\theta_{l}}(z^{*})$ from the latent MLP is utilized to assist the render MLP in rendering.
However, directly using this output $f_{\theta_{l}}(z^{*})$ as the input of the render MLP is akin to using the latent code itself, as demonstrated in Eq. \ref{con:liif}, resulting in no efficiency improvement yet:
\begin{equation}
  I(x^{*}_{i, j}) = f_{\theta_{r}}(f_{\theta_{l}}(z^{*}), x^{*}_{i, j}).
  \label{con:c2f}
\end{equation}

Furthermore, to render arbitrary pixels accurately, it is necessary for the render MLP to operate in HR space. Therefore, the most viable approach to minimize rendering computations is to reduce the dimension or depth of the render MLP. However, an HD render MLP is still needed in Eq. \ref{con:c2f} to fully receive and process the output $f_{\theta_{l}}(z^{*})$, which contradicts our intention of utilizing an LD render MLP. To overcome these limitations, modulations become a suitable solution to connect the latent and render space. Modulations have been used in INR \cite{skorokhodov2021adversarial,mehta2021modulated,dupont2022coin++} and are often associated with the hyper-base network design. A hyper network converts an input image to modulations, which are then utilized by a small base network to map coordinates to a continuous image. However, these modulations are shared by the input image and work at the image level, resulting in unsatisfactory performance and challenging training.
Hence, we introduce a latent modulation $m^{*} \in \mathbb{R}^{D_{M}}$, which is generated by the latent MLP using the latent code $z^{*}$. The latent modulation consists of the FiLM layers \cite{perez2018film}, where each layer consists of a scale modulation and a shift modulation:
\begin{equation}
  m^{*} = [\alpha^{*}_{1}, \beta^{*}_{1}, \dots, \alpha^{*}_{K}, \beta^{*}_{K}] = Split(f_{\theta_{l}}(z^{*})),
  \label{con:latent-modulation}
\end{equation}
where $\alpha^{*}$ represents the element-wise scale modulation, $\beta^{*}$ represents the element-wise shift modulation, and $K$ represents the depth of hidden layers in the render MLP.

For each HR coordinate, the latent modulation $m^{*}$ of the latent code $z^{*}$ nearest to this coordinate is applied to the render MLP, and is applied to the hidden features $h^{*} \in \mathbb{R}^{D_{H}}$ after each hidden linear layer:
\begin{equation}
  h^{*}_{k+1} = f_{\theta^{k+1}_{r}}(\sigma((1 + \alpha^{*}_{k}) \odot h^{*}_{k} + \beta^{*}_{k})), \\
  \label{con:modulation-process}
\end{equation}
where $h^{*}_{k}$ represents the hidden features after the $k$-th hidden linear layer, and $\sigma$ represents the activation function.

Importantly, the proposed latent modulation effectively solves the incompatibility between LR-HD latent codes and HR-LD rendering and significantly disentangles the computations from parameters. For a latent modulation with 192 dimensions, we can accurately render images using a 7-layer render MLP with only 16 dimensions.
This LD render MLP only learns the general rendering rules of natural images, while the latent modulation optimizes its rendering process based on the characteristics of each input latent code. As a result, a tiny render MLP can quickly adapt to any input latent code and accomplish efficient rendering at arbitrary resolution. A remaining concern is that the latent codes in existing INR-based ASSR methods often possess high dimensions, resulting in unnecessary computations when directly input into the render MLP. Therefore, we compress the HD latent code to an LD one $z^{*}_{c}$ using the latent MLP, extending Eq. \ref{con:latent-modulation} to the first stage in Eq. \ref{con:lmf}. Building on the latent modulation, we propose a computational optimal paradigm of continuous image representation, called Latent Modulated Function (LMF):
\begin{equation}
\begin{split}
    \text{Stage 1: }& [m^{*}, z^{*}_{c}]= Split(f_{\theta_{l}}(z^{*})), \\
    \text{Stage 2: }& I(x^{*}_{i, j}) = f_{\theta_{r}, m^{*}}(z^{*}_{c}, x^{*}_{i, j}).
  \label{con:lmf}
\end{split}
\end{equation}

LMF is a universal paradigm for efficient ASSR and is compatible with existing INR-based ASSR methods. By effortlessly relocating the time-consuming operations and large MLPs to latent space, LMF allows a minimal render MLP to perform the same level of ASSR as the original ASSR method. In addition, LMF effectively decouples the computational expensive techniques of feature unfolding and local ensemble \cite{chen2021learning}. Specifically, LMF only utilizes feature unfolding for the latent MLP, where nearby $3\times3$ feature vectors are concatenated as inputs. LMF performs local ensemble exclusively for the render MLP, which merges the predictions from the top-left, top-right, bottom-left, and bottom-right latent codes to achieve continuous predictions.

\begin{figure}[t]
  \centering
  \includegraphics[width=0.82\linewidth]{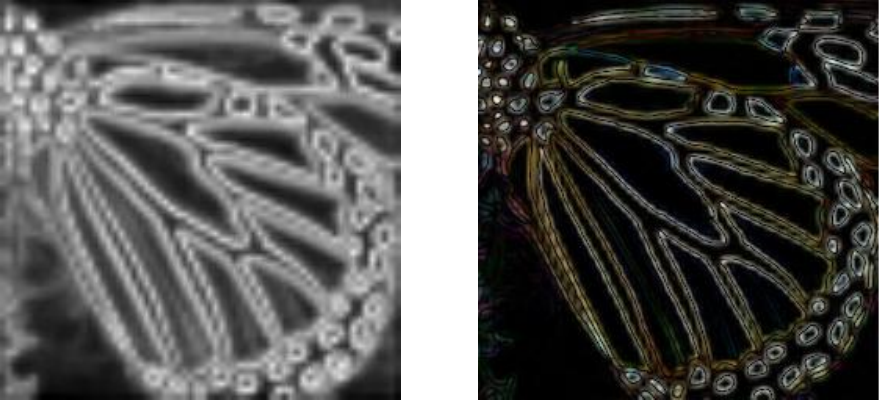}
  \caption{Visualization ($\times4$ SR) of the normalized mean of shift modulation (left), and the normalized residual between the LM-LIIF predicted image and the bilinear upsampled image (right).}
  \label{fig:mod-mean-visual}
  \vspace{-0.45cm}
\end{figure}

\begin{figure}[t]
  \centering
  \includegraphics[width=0.86\linewidth]{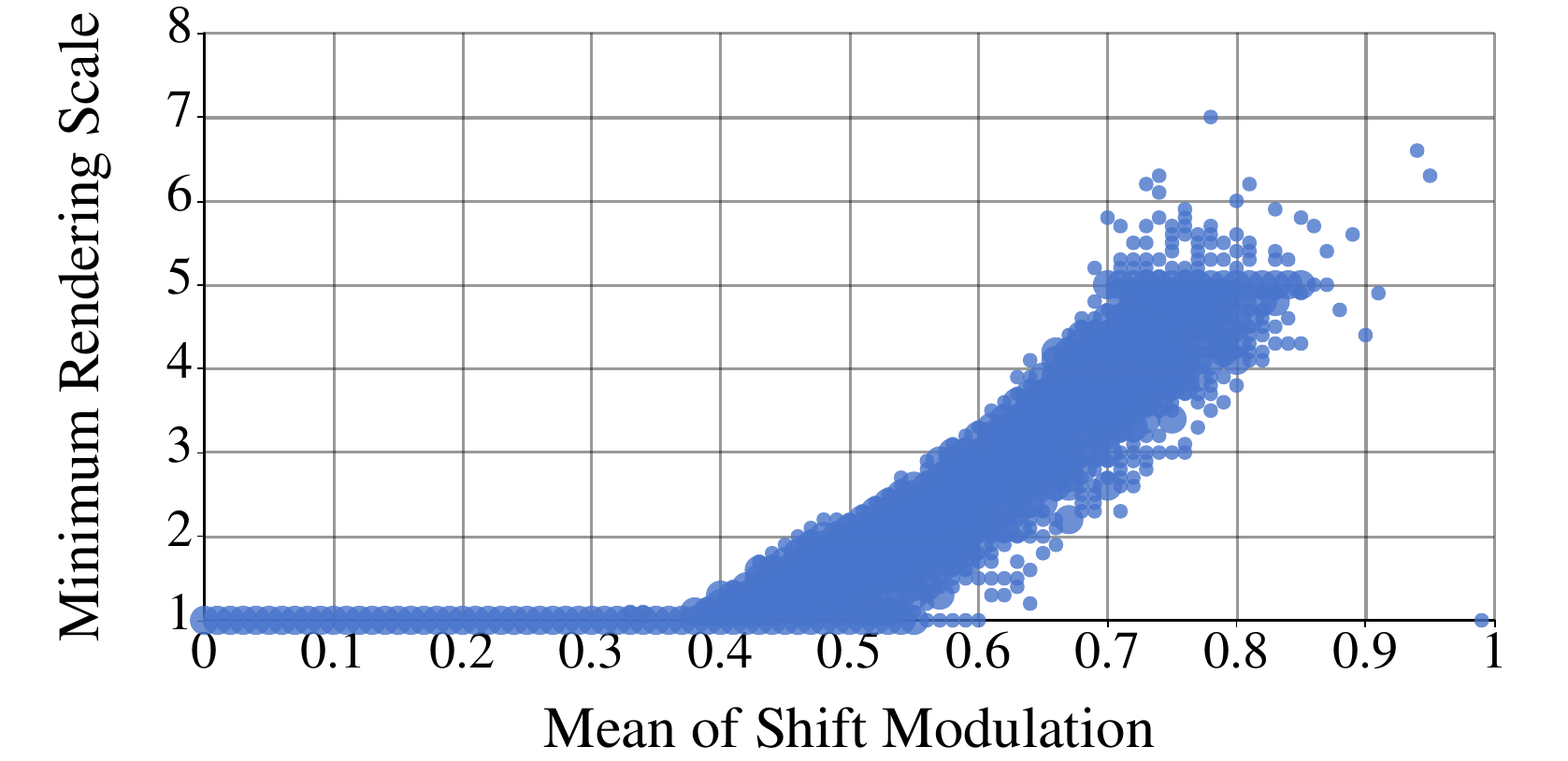}
  \caption{The positive correlation between the means of shift modulation and the minimum rendering scale factors in LM-LIIF.}
  \label{fig:mod-scale-relation}
  \vspace{-0.45cm}
\end{figure}

\subsection{Controllable Multi-Scale Rendering}
The HR regions of two latent codes often have different signal complexity, allowing for the use of different decoding functions. However, the challenge lies in obtaining the signal complexity of each latent code.
The modulation scheme in Eq. \ref{con:modulation-process} suggests that the intensity of latent modulation is proportional to the signal complexity of its latent code. A lower intensity implies that the signal can be sufficiently rendered by the render MLP itself.
In Figure \ref{fig:mod-mean-visual}, we display the normalized mean of shift modulation (left), and the normalized residual between the SR result of LMF-based LIIF and the bilinear upsampled result (right). It is evident that the means of modulation closely align with the signal complexity of the predicted image. Regions with higher means of modulation consistently correspond to more complex textures and edges.
Hence, we employ the mean of shift modulation as the intensity of latent modulation.

In LMF, we define the signal complexity of the latent code $z^{*}$ as the minimum scale factor $s^{*}$ needed to fully represent its signal. In simple terms, any rendering with a scale factor $s, s>s^{*}$ will yield the same result as rendering with $s^{*}$ followed by upsample interpolation (\ie bilinear) with scale $\frac{s}{s^{*}}$. Therefore, we only need to render $z^{*}$ with scale $min(s^{*}, s)$ when upsampling an image with a scale $s$.
To obtain the minimum scale factor of each mean of shift modulation, in Figure \ref{fig:mod-scale-relation}, we sample 100 means of shift modulation uniformly from the range of $[0, 1]$ and measure their minimum scale factors on the LR images in the training set.
Figure \ref{fig:mod-scale-relation} clearly shows the positive correlation between the means of shift modulation and the minimum scale factors. However, since they do not have a one-to-one correspondence, we create a Scale2Mods table by sampling common scale factors and recording their minimum and maximum means of shift modulation.

\begin{table*}[ht]
\centering
\footnotesize
\caption{Quantitative comparison (PSNR (dB)/MACs) for ASSR on DIV2K validation set. The best results are highlighted in \textbf{bold}.}
\label{tab:div2k100_psnr_result}
\begin{tabular}{l|c|ccc|ccc}
\hline
\multirow{2}{*}{Method} & \multirow{2}{*}{Params} & \multicolumn{3}{c|}{In-scale} & \multicolumn{3}{c}{Out-of-scale} \bigstrut[t]\\

& & $\times2$ & $\times3$ & $\times4$ & $\times6$ & $\times12$ & $\times18$ \bigstrut[t]\\
\hline

Bicubic & - & 31.01/- & 28.22/- & 26.66/- & 24.82/- & 22.27/- & 21.00/-\bigstrut[t]\\

EDSR-b \cite{lim2017enhanced} & - & 34.55/- & 30.90/- & 28.94/- & - & - & - \bigstrut[t]\\
\cline{2-8}

LIIF \cite{chen2021learning} & 355K & \textbf{34.67}/3.567T & \textbf{30.96}/3.567T & \textbf{29.00}/3.567T & \textbf{26.75}/3.567T & 23.71/3.567T & 22.17/3.567T \bigstrut[t]\\

LM-LIIF & \textbf{170K} & 34.65/\textbf{125.32G} & 30.95/\textbf{65.37G} & 28.99/\textbf{44.38G} & \textbf{26.75}/\textbf{25.30G} & \textbf{23.72/6.61G} & \textbf{22.19/2.95G} \bigstrut[t]\\ 
\cline{2-8}

LTE \cite{lee2022local} & 506K & \textbf{34.72}/2.228T & \textbf{31.02}/2.123T & \textbf{29.04}/2.086T & \textbf{26.81}/2.059T & \textbf{23.78}/2.044T & \textbf{22.23}/2.026T \bigstrut[t]\\

LM-LTE & \textbf{278K} & 34.71/\textbf{198.81G} & 30.99/\textbf{100.75G} & 29.03/\textbf{66.43G} & 26.78/\textbf{34.96G} & 23.74/\textbf{9.07G} & 22.21/\textbf{4.10G} \bigstrut[t]\\
\cline{2-8}

CiaoSR-1.4M \cite{cao2023ciaosr} & 1.429M & \textbf{34.91}/11.460T & \textbf{31.15}/11.443T & \textbf{29.23}/11.437T & \textbf{26.95}/11.433T & \textbf{23.88}/11.430T & \textbf{22.32}/11.348T \bigstrut[t]\\

LM-CiaoSR & \textbf{753K} & 34.81/\textbf{1.071T} & 31.10/\textbf{501.36G} & 29.12/\textbf{293.01G} & 26.86/\textbf{135.66G} & 23.80/\textbf{42.60G} & 22.26/\textbf{23.90G} \bigstrut[t]\\
\hline
\end{tabular}
\end{table*}

\begin{table*}[ht]
\centering
\caption{Quantitative comparison (PSNR (dB)) for in-scale SR on test sets. The best results are highlighted in \textbf{bold}.
}
\label{tab:benchmark_result_in_scale}
\resizebox{\linewidth}{!}{
\begin{tabular}{l|ccc|ccc|ccc|ccc|ccc}
\hline
\multirow{2}{*}{Method} & \multicolumn{3}{c|}{Set5} & \multicolumn{3}{c|}{Set14} & \multicolumn{3}{c|}{BSD100} & \multicolumn{3}{c|}{Urban100}& \multicolumn{3}{c}{DIV2K100}\bigstrut[t]\\

& $\times2$ & $\times3$ & $\times4$
& $\times2$ & $\times3$ & $\times4$ 
& $\times2$ & $\times3$ & $\times4$
& $\times2$ & $\times3$ & $\times4$
& $\times2$ & $\times3$ & $\times4$ \bigstrut[t]\\
\hline

RDN \cite{zhang2018residual} 
& 38.24 & 34.71 & 32.47
& 34.01 & 30.57 & 28.81
& 32.34 & 29.26 & 27.72
& 32.89 & 28.80 & 26.61
& 34.94 & 31.22 & 29.19
\bigstrut[t]\\
\cline{2-16}

LIIF \cite{chen2021learning} 
& 38.17 & 34.68 & \textbf{32.50} 
& 33.97 & 30.53 & 28.80 
& \textbf{32.32} & 29.26 & 27.74 
& \textbf{32.87} & \textbf{28.82} & \textbf{26.68}
& \textbf{34.99} & \textbf{31.26} & \textbf{29.27}
\bigstrut[t]\\
LM-LIIF
& \textbf{38.18} & \textbf{34.72} & \textbf{32.50}
& \textbf{34.06} & \textbf{30.55} & \textbf{28.83}
& \textbf{32.32} & \textbf{29.28} & \textbf{27.75}
& 32.82 & 28.81 & 26.66
& 34.96 & 31.25 & \textbf{29.27}
\bigstrut[t]\\
\cline{2-16}
LTE \cite{lee2022local}
& \textbf{38.23} & 34.72 & \textbf{32.61}
& 34.09 & \textbf{30.58} & \textbf{28.88}
& 32.36 & 29.30 & 27.77
& \textbf{33.04} & \textbf{28.97} & \textbf{26.81}
& \textbf{35.04} & \textbf{31.32} & \textbf{29.33}
\bigstrut[t]\\
LM-LTE
& \textbf{38.23} & \textbf{34.76} & 32.53
& \textbf{34.11} & 30.56 & 28.86
& \textbf{32.37} & \textbf{29.31} & \textbf{27.78}
& 33.03 & 28.96 & 26.80
& 35.03 & 31.31 & 29.32
\bigstrut[t]\\
\cline{2-16}

CiaoSR-1.4M \cite{cao2023ciaosr}
& \textbf{38.29} & \textbf{34.85} & \textbf{32.66} 
& \textbf{34.22} & \textbf{30.65} & \textbf{28.93} 
& \textbf{32.41} & \textbf{29.34} & \textbf{27.83} 
& \textbf{33.30} & \textbf{29.17} & \textbf{27.11}
& \textbf{35.15} & \textbf{31.42} & \textbf{29.45}
\bigstrut[t]\\
LM-CiaoSR
& 38.24 & 34.81 & 32.59
& 34.11 & 30.62 & 28.90
& 32.39 & 29.32 & 27.80
& 33.13 & 29.13 & 26.95
& 35.08 & 31.36 & 29.37
\bigstrut[t]\\
\hline

SwinIR \cite{liang2021swinir} 
& 38.35 & 34.89 & 32.72
& 34.14 & 30.77 & 28.94 
& 32.44 & 29.37 & 27.83 
& 33.40 & 29.29 & 27.07
& 35.26 & 31.50 & 29.48
\bigstrut[t]\\
\cline{2-16}

LIIF \cite{chen2021learning} 
& 38.28 & 34.87 & 32.73
& 34.14 & 30.75 & 28.98
& 32.39 & 29.34 & 27.84
& 33.36 & 29.33 & 27.15
& 35.17 & 31.46 & 29.46
\bigstrut[t]\\
LM-LIIF
& \textbf{38.31} & \textbf{34.88} & \textbf{32.78}
& \textbf{34.31} & \textbf{30.79} & \textbf{29.02}
& \textbf{32.43} & \textbf{29.37} & \textbf{27.85}
& \textbf{33.46} & \textbf{29.43} & \textbf{27.21}
& \textbf{35.21} & \textbf{31.48} & \textbf{29.49}
\bigstrut[t]\\
\cline{2-16}

LTE \cite{lee2022local} 
& \textbf{38.33} & \textbf{34.89} & \textbf{32.81}
& 34.25 & \textbf{30.80} & \textbf{29.06}
& 32.44 & \textbf{29.39} & 27.86
& 33.50 & 29.41 & \textbf{27.24}
& \textbf{35.24} & \textbf{31.50} & \textbf{29.51}
\bigstrut[t]\\
LM-LTE
& 38.32 & 34.88 & 32.77
& \textbf{34.28} & 30.79 & 29.01
& \textbf{32.46} & \textbf{29.39} & \textbf{27.87}
& \textbf{33.52} & \textbf{29.44} & \textbf{27.24}
& \textbf{35.24} & \textbf{31.50} & 29.50
\bigstrut[t]\\
\cline{2-16}

CiaoSR-3.1M \cite{cao2023ciaosr}
& \textbf{38.38} & 34.91 & 32.84
& \textbf{34.33} & 30.82 & \textbf{29.08}
& 32.47 & 29.42 & \textbf{27.90}
& \textbf{33.65} & 29.52 & \textbf{27.42}
& 35.29 & \textbf{31.55} & \textbf{29.59}
\bigstrut[t]\\
LM-CiaoSR
& 38.37 & \textbf{34.94} & \textbf{32.87}
& \textbf{34.33} & \textbf{30.83} & \textbf{29.08}
& \textbf{32.48} & \textbf{29.43} & \textbf{27.90}
& \textbf{33.65} & \textbf{29.58} & 27.39
& \textbf{35.30} & \textbf{31.55} & 29.56
\bigstrut[t]\\
\hline
\end{tabular}
}
\vspace{-0.4cm}
\end{table*}

\begin{table*}[ht]
\centering
\caption{Quantitative comparison (PSNR (dB)) for out-of-scale SR on test sets. The best results are highlighted in \textbf{bold}. The first 6 rows use RDN as the encoder, and the last 6 rows use SwinIR as the encoder.
}
\label{tab:benchmark_result_out_of_scale}
\resizebox{\linewidth}{!}{
\begin{tabular}{l|ccc|ccc|ccc|ccc|ccc}
\hline
\multirow{2}{*}{Method} & \multicolumn{3}{c|}{Set5} & \multicolumn{3}{c|}{Set14} & \multicolumn{3}{c|}{BSD100} & \multicolumn{3}{c|}{Urban100} & \multicolumn{3}{c}{DIV2K100}\bigstrut[t]\\

& $\times6$ & $\times8$ & $\times12$
& $\times6$ & $\times8$ & $\times12$
& $\times6$ & $\times8$ & $\times12$
& $\times6$ & $\times8$ & $\times12$
& $\times6$ & $\times12$ & $\times18$\bigstrut[t]\\
\hline

LIIF \cite{chen2021learning} 
& 29.15 & 27.14 & \textbf{24.86} 
& \textbf{26.64} & \textbf{25.15} & 23.24 
& \textbf{25.98} & 24.91 & 23.57 
& \textbf{24.20} & \textbf{22.79} & \textbf{21.15}
& \textbf{26.99} & 23.89 & 22.34
\bigstrut[t]\\
LM-LIIF
& \textbf{29.18} & \textbf{27.21} & 24.71
& \textbf{26.66} & \textbf{25.17} & \textbf{23.31}
& \textbf{25.98} & \textbf{24.93} & \textbf{23.58}
& 24.14 & 22.77 & \textbf{21.15}
& \textbf{26.99} & \textbf{23.91} & \textbf{22.36}
\bigstrut[t]\\
\cline{2-16}

LTE \cite{lee2022local}
& \textbf{29.32} & \textbf{27.26} & 24.79 
& \textbf{26.71} & \textbf{25.16} & 23.31 
& \textbf{26.01} & \textbf{24.95} & 23.60 
& \textbf{24.28} & \textbf{22.88} & 21.22
& \textbf{27.04} & \textbf{23.95} & \textbf{22.40}
\bigstrut[t]\\
LM-LTE
& 29.21 & 27.22 & \textbf{24.82}
& 26.69 & 25.15 & \textbf{23.32}
& \textbf{26.01} & \textbf{24.95} & \textbf{23.61}
& 24.26 & 22.86 & \textbf{21.24}
& 27.03 & \textbf{23.95} & 22.39
\bigstrut[t]\\
\cline{2-16}

CiaoSR-1.4M \cite{cao2023ciaosr}
& \textbf{29.46} & \textbf{27.36} & \textbf{24.92}
& \textbf{26.79} & \textbf{25.28} & \textbf{23.37}
& \textbf{26.07} & \textbf{25.00} & \textbf{23.64} 
& \textbf{24.58} & \textbf{23.13} & \textbf{21.42}
& \textbf{27.16} & \textbf{24.06} & \textbf{22.48}
\bigstrut[t]\\
LM-CiaoSR
& 29.24 & 27.26 & 24.80
& 26.72 & 25.18 & 23.30
& 26.03 & 24.81 & 23.60
& 24.40 & 22.98 & 21.33
& 27.07 & 23.98 & 22.42
\bigstrut[t]\\                                           
\hline

LIIF \cite{chen2021learning} 
& \textbf{29.46} & \textbf{27.36} & \textbf{24.99}
& \textbf{26.82} & 25.34 & 23.39
& 26.07 & 25.01 & \textbf{23.65}
& 24.59 & 23.14 & 21.43
& 27.15 & 24.02 & 22.43
\bigstrut[t]\\
LM-LIIF
& \textbf{29.46} & \textbf{27.36} & 24.98
& \textbf{26.82} & \textbf{25.37} & \textbf{23.43}
& \textbf{26.08} & \textbf{25.03} & \textbf{23.65}
& \textbf{24.61} & \textbf{23.15} & \textbf{21.49}
& \textbf{27.18} & \textbf{24.06} & \textbf{22.49}
\bigstrut[t]\\
\cline{2-16}

LTE \cite{lee2022local} 
& 29.50 & 27.35 & \textbf{25.07}
& \textbf{26.86} & \textbf{25.42} & \textbf{23.44}
& 26.09 & \textbf{25.03} & \textbf{23.66}
& 24.62 & \textbf{23.17} & \textbf{21.50}
& \textbf{27.20} & \textbf{24.09} & \textbf{22.50}
\bigstrut[t]\\
LM-LTE
& \textbf{29.52} & \textbf{27.39} & 25.01
& 26.83 & 25.38 & 23.41
& \textbf{26.10} & \textbf{25.03} & \textbf{23.66}
& \textbf{24.63} & \textbf{23.17} & \textbf{21.50}
& 27.19 & 24.08 & \textbf{22.50}
\bigstrut[t]\\
\cline{2-16}

CiaoSR-3.1M \cite{cao2023ciaosr}
& \textbf{29.62} & \textbf{27.45} & 24.96 
& \textbf{26.88} & \textbf{25.42} & 23.38 
& \textbf{26.13} & \textbf{25.07} & \textbf{23.68} 
& \textbf{24.84} & \textbf{23.34} & \textbf{21.60}
& \textbf{27.28} & \textbf{24.15} & \textbf{22.54}
\bigstrut[t]\\
LM-CiaoSR
& 29.52 & 27.42 & \textbf{25.02}
& 26.84 & 25.41 & \textbf{23.45}
& 26.12 & 25.06 & \textbf{23.68}
& 24.74 & 23.26 & 21.58
& 27.24 & 24.12 & 22.53
\bigstrut[t]\\
\hline
\end{tabular}
}
\end{table*}

\begin{table*}[ht]
\centering
\footnotesize
\caption{Efficiency comparison (MACs/GPU runtime) for ASSR. The best results are highlighted in \textbf{bold}.
}
\label{tab:efficiency_comparison}
\resizebox{\linewidth}{!}{
\begin{tabular}{l|c|c|ccc|ccc|ccc}
\hline
\multirow{2}{*}{Decoder} & \multirow{2}{*}{Params} & GPU & \multicolumn{3}{c|}{180p} & \multicolumn{3}{c|}{270p} & \multicolumn{3}{c}{360p} \bigstrut[t]\\
 & & mem. & 720p & 1080p & 2k & 720p & 1080p & 2k & 720p & 1080p & 2k \bigstrut[t]\\
\hline

LIIF \cite{chen2021learning} & 355K & 7.7GB & \makecell{1.160T/\\1.1148s} & \makecell{2.609T/\\2.5230s} & \makecell{4.638T/\\4.5232s} & \makecell{1.160T/\\1.1195s} & \makecell{2.609T/\\2.5347s} & \makecell{4.638T/\\4.5544s} & \makecell{1.160T/\\1.1272s} & \makecell{2.609T/\\2.5456s} & \makecell{4.638T/\\4.5589s} \bigstrut[t]\\

LM-LIIF & \textbf{170K} & \textbf{5.3GB} & \makecell{\textbf{14.43G}/\\ \textbf{0.0246s}} & \makecell{\textbf{18.72G}/\\ \textbf{0.0474s}} & \makecell{\textbf{24.72G}/\\ \textbf{0.0795s}} & \makecell{\textbf{25.39G}/\\ \textbf{0.0363s}} & \makecell{\textbf{32.46G}/\\ \textbf{0.0593s}} & \makecell{\textbf{\iffalse 42.36G \fi 37.50G}/\\ \textbf{0.0919s}} & \makecell{\textbf{40.74G/}\\ \textbf{0.0541s}} & \makecell{\textbf{47.81G/}\\ \textbf{0.0770s}} & \makecell{\textbf{57.71G/}\\ \textbf{0.1107s}} \bigstrut[t]\\
\cline{2-12}

LTE \cite{lee2022local} & 506K & 8.4GB & \makecell{694.34G/\\0.9135s} & \makecell{1.506T/\\2.0366s} & \makecell{2.666T/\\3.7578s} & \makecell{714.11G/\\0.9191s} & \makecell{1.526T/\\2.0497s} & \makecell{2.685T/\\3.7694s} & \makecell{741.80G/\\0.9258s} & \makecell{1.553T/\\2.0637s} & \makecell{2.712T/\\3.7791s} \bigstrut[t]\\

LM-LTE & \textbf{278K} & \textbf{4.7GB} & \makecell{\textbf{21.60G}/\\ \textbf{0.0382s}} & \makecell{\textbf{25.87G}/\\ \textbf{0.0750s}} & \makecell{\textbf{33.21G}/\\ \textbf{0.1254s}} & \makecell{\textbf{39.53G}/\\ \textbf{0.0545s}} & \makecell{\textbf{48.59G}/\\ \textbf{0.0905s}} & \makecell{\textbf{\iffalse 61.28 \fi 52.88G}/\\ \textbf{0.1414s}} & \makecell{\textbf{64.63G/}\\ \textbf{0.0774s}} & \makecell{\textbf{73.69G/}\\ \textbf{0.1123s}} & \makecell{\textbf{86.38G/}\\ \textbf{0.1640s}} \bigstrut[t]\\
\cline{2-12}

CiaoSR-1.4M \cite{cao2023ciaosr} & 1.429M & \textbf{8.8GB} & \makecell{3.718T/\\1.8820s} & \makecell{8.362T/\\4.1718s} & \makecell{14.864T/\\7.3619s} & \makecell{3.721T/\\1.9838s} & \makecell{8.365T/\\4.2999s} & \makecell{14.867T/\\7.5611s} & \makecell{3.725T/\\2.1355s} & \makecell{8.370T/\\4.5064s} & \makecell{14.871T/\\7.8366s}\bigstrut[t]\\

LM-CiaoSR & \textbf{753K} & 8.9GB & \makecell{\textbf{95.23G}/\\ \textbf{0.2637s}} & \makecell{\textbf{99.57G}/\\ \textbf{0.3849s}} & \makecell{\textbf{106.68G}/\\ \textbf{0.5546s}} & \makecell{\textbf{204.09G}/\\ \textbf{0.5263s}} & \makecell{\textbf{214.32G}/\\ \textbf{0.6476s}} & \makecell{\textbf{\iffalse 228.64G \fi 217.83G}/\\ \textbf{0.8160s}} & \makecell{\textbf{356.46G/}\\ \textbf{0.8844s}} & \makecell{\textbf{366.69G/}\\ \textbf{1.0013s}} & \makecell{\textbf{381.02G/}\\ \textbf{1.1778s}} \bigstrut[t]\\

\hline
\end{tabular}
}
\vspace{-0.4cm}
\end{table*}

\begin{figure*}[ht]
  \centering
  \includegraphics[width=0.98\textwidth]{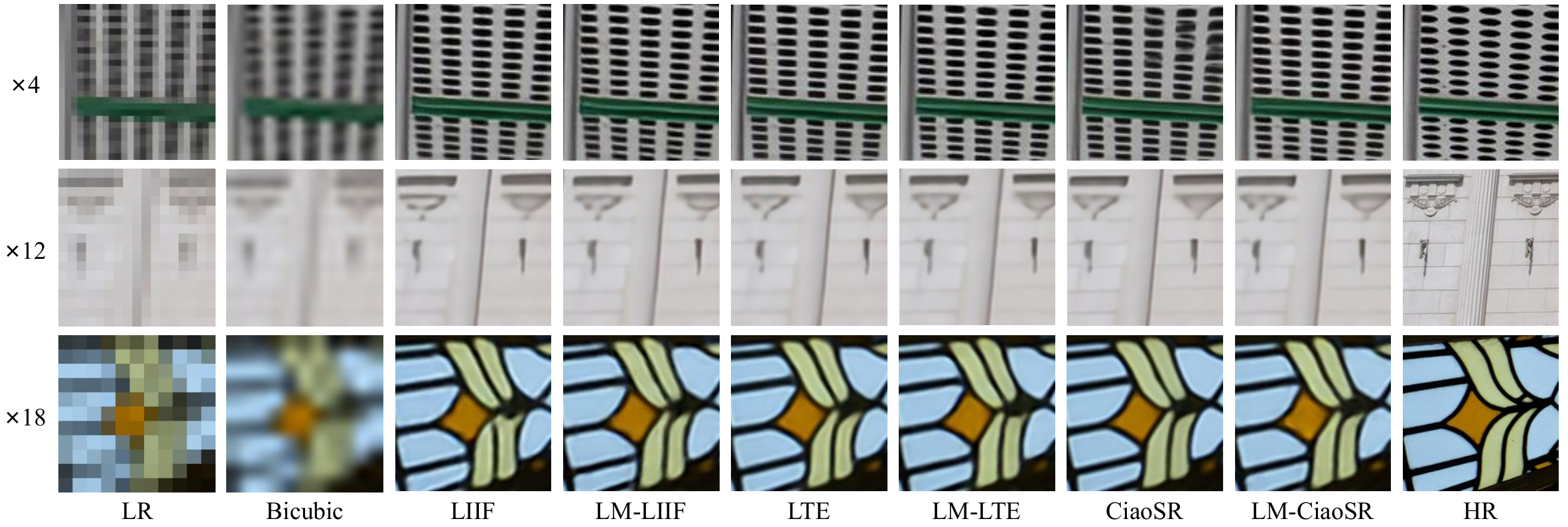}
  \caption{Qualitative comparison for ASSR. All of the ASSR methods use SwinIR as the encoder.}
  \label{fig:qualitative_comparison}
\end{figure*}

\begin{table*}[ht]
\centering
\footnotesize
\caption{Ablation study (PSNR(dB)/MACs on DIV2K validation set) on latent modulations and latent codes, using EDSR-b as the encoder.
}
\label{tab:ablation_study_modulations}
\begin{tabular}{l|c|c|c|c|c|cc|cc}
\hline
\multirow{2}{*}{Decoder} & Latent & Latent & Latent & Render & \multirow{2}{*}{Params} & \multicolumn{2}{c|}{In-scale} & \multicolumn{2}{c}{Out-of-scale} \bigstrut[t]\\
& code & modulation & MLP & MLP & & $\times2$ & $\times4$ & $\times8$ & $\times16$ \bigstrut[t]\\
\hline

LIIF & 576 & $\times$ & $\times$ & $5\times256$ & 355K & \textbf{34.67}/3.567T & \textbf{29.00}/3.567T & \textbf{25.40}/3.567T & 22.61/3.567T \bigstrut[t]\\
LIIF (tiny) & 576 & $\times$ & $\times$ & $9\times16$ & \textbf{12K} & 33.61/117.44G & 28.18/117.44G  & 24.78/117.44G & 22.16/117.44G \bigstrut[t]\\
LIIF (C2F) & 192 & $\times$ & $2\times192$ & $7\times16$ & 202K & 34.11/221.54G & 28.63/114.62G & 25.11/73.04G & 22.38/58.91G \bigstrut[t]\\
LM-LIIF & 64 & Scale & $2\times192$ & $7\times16$ & 136K & 34.61/\textbf{110.94G} & 28.96/46.87G & 25.38/30.79G & 22.61/26.58G \bigstrut[t]\\
LM-LIIF & 64 & Shift & $2\times192$ & $7\times16$ & 136K & 34.59/\textbf{110.94G} & 28.94/46.87G & 25.36/30.79G & 22.59/26.58G \bigstrut[t]\\
LM-LIIF & 64 & Scale \& shift & $2\times192$ & $7\times16$ & 155K & 34.66/123.10G & \textbf{29.00}/49.91G & \textbf{25.40}/31.55G & \textbf{22.63}/26.77G \bigstrut[t]\\
LM-LIIF & 16 & Scale \& shift & $2\times208$ & $7\times16$ & 170K & 34.65/125.32G & 28.99/\textbf{44.38G} & \textbf{25.40}/\textbf{24.10G} & 22.62/\textbf{18.90G} \bigstrut[t]\\
\hline
\end{tabular}
\vspace{-0.4cm}
\end{table*}

Based on the Scale2Mods table, we propose a Controllable Multi-Scale Rendering (CMSR) algorithm to decode latent codes with their minimum scale factors.
Firstly, we initialize $s_{-1}$ as 1, $I_{s_{-1}}$ as the LR image. For each sampled scale $s_{i} \in [s_{0}, s_{1}, \dots, s]$, we first upsample the last result $I_{s_{i-1}}$ using bilinear interpolation with $\frac{s_{i}}{s_{i-1}}$ to obtain $I_{s_{i}}$. We then query the Scale2Mods table to obtain the range of modulation means $[m^{s_{i}}_{min}, m^{s_{i}}_{max}]$ for $s_{i}$. For $s_{i}$, the render MLP only needs to render the latent codes whose corresponding modulation means falling within this range. We repeat the above process iteratively until we obtain the final SR result or until all latent codes have been decoded.
For precision controllable rendering, when creating the Scale2Mods table, we set an MSE threshold $t$ and define the signal complexity of a latent code as the minimum scale factor needed to render its signal within this error $t$.
By effortlessly adjusting the MSE threshold during testing, CMSR is capable of rendering with varying precision and speed, while adapting naturally to the image content.

\section{Experiments}
\subsection{Experimental Setups}
\noindent \textbf{Dataset.} Following the setups in previous ASSR methods \cite{chen2021learning, lee2022local, cao2023ciaosr}, we employ the DIV2K dataset \cite{agustsson2017ntire} as the training set.
In each training iteration, we uniformly sample a downsampling scale $s$ from a continuous range of $[\times1, \times4]$. HR patches with spatial size of $48s\times48s$ are cropped and bicubic downsampled to LR patches with spatial size of $48\times48$. To supervise the training process, we randomly sample $48\times48$ pixels from each HR patch.
For evaluation, we present results on the DIV2K validation set (DIV2K 100) along with four widely-used benchmark test sets: Set5 \cite{bevilacqua2012low}, Set14 \cite{zeyde2012single}, BSD100 \cite{martin2001database}, and Urban100 \cite{huang2015single}.

\noindent \textbf{Implementation Details.} We transform three previous ASSR methods, namely LIIF \cite{chen2021learning}, LTE \cite{lee2022local}, and CiaoSR \cite{cao2023ciaosr}, into our proposed LMF. LM-LIIF employs a 2-layer latent MLP with 208 dimensions, and a 7-layer render MLP with 16 dimensions. As for LM-LTE, we employ a 2-layer latent MLP with 256 dimensions, a 9-layer render MLP with 16 dimensions, and a local texture estimator \cite{lee2022local} with 128 dimensions. LM-CiaoSR employs four 2-layer latent MLPs with 256 dimensions to generate key, value, and query information, followed by a 9-layer render MLP with 16 dimensions. The detailed architectures of LMF-based methods are provided in the supplementary material.

We utilize EDSR-baseline (EDSR-b) \cite{lim2017enhanced}, RDN \cite{zhang2018residual}, and SwinIR \cite{liang2021swinir} without their upsampling modules as encoders. For both EDSR-b and RDN encoders, we conduct training for $1000$ epochs with a batch size of $16$. The learning rate is initialized as $1 \times 10^{-4}$ and decayed by a factor of $0.5$ every $200$ epochs. As for the SwinIR encoder, we conduct training for $1000$ epochs with a batch size of $32$. The learning rate is initialized as $2 \times 10^{-4}$ and decayed by a factor of $0.5$ at $[500, 800, 900, 950]$ epochs. For optimization, we utilize the L1 pixel loss as the loss function and employ the Adam \cite{kingma2015adam} optimizer with $\beta_{1} = 0.9$, $\beta_{2} = 0.999$. Our training details align with previous ASSR works \cite{lee2022local, cao2023ciaosr}.

\noindent \textbf{Evaluation Settings.} In line with previous methods \cite{chen2021learning, lee2022local, cao2023ciaosr}, we employ PSNR as the objective criterion to assess quality performance.
To evaluate efficiency, we measure the computational cost using Multiply-Accumulate Operations (MACs) and quantify the model speed using the runtime on an RTX 2080 Ti GPU. Additionally, we provide the number of parameters and GPU memory usage. 
Notably, the MACs, runtime, GPU memory, and parameters are evaluated solely on the decoder unless explicitly stated. 

\subsection{ASSR Comparison}
\noindent \textbf{Quantitative Results.}
In Table \ref{tab:div2k100_psnr_result}, we compare our LMF-based methods with reference ASSR methods, namely LIIF \cite{chen2021learning}, LTE \cite{lee2022local}, and CiaoSR \cite{cao2023ciaosr}, on the DIV2K validation set using EDSR-b as the encoder. We include the results of bicubic interpolation and EDSR-b with its initial upsampling module for reference. The results demonstrate that LMF-based methods achieve a significant reduction in computational cost while delivering competitive PSNR performance compared to the original methods. Compared to LIIF, LM-LIIF achieves a reduction of $96.6\%$ to $99.9\%$ in computational cost across all scales. Similarly, LM-LTE and LM-CiaoSR provide $91.3\%$ to $99.8\%$ and $90.4\%$ to $99.8\%$ reduction in computational cost compared to LTE and CiaoSR, respectively. However, LM-CiaoSR exhibits slightly lower PSNR than CiaoSR due to relocating the expensive attention-in-attention modules from HR-HD space to latent space for the minimal computational cost.

In Table \ref{tab:benchmark_result_in_scale} and \ref{tab:benchmark_result_out_of_scale}, we present the comprehensive PSNR evaluation for ASSR using RDN and SwinIR as the encoders. Notably, CiaoSR utilizes a 1.429M decoder for EDSR-b and RDN, and a 3.065M decoder for SwinIR, while our LM-CiaoSR employs the same decoder for all encoders.
The results demonstrate the competitive PSNR performance of our LMF-based methods, especially LM-LIIF and LM-LTE, compared to the reference methods on benchmark test sets. Notably, LMF-based methods exhibit superior PSNR performance gains on the SwinIR encoder. The results presented in Table \ref{tab:div2k100_psnr_result}, \ref{tab:benchmark_result_in_scale}, and \ref{tab:benchmark_result_out_of_scale} collectively validate the universal applicability of our LMF across different encoders, INR-based ASSR methods, test sets, and scales.

In Table \ref{tab:efficiency_comparison}, we present comparisons of MACs, runtime, GPU memory, and parameters for ASSR across multiple input and output resolutions. We utilize bicubic downsampled images from the DIV2K validation set as inputs for each resolution, and all results exclude the encoders. As shown in the table, our LMF-based methods exhibit a significant reduction in computational cost, ranging from $90.7\%$ to $99.5\%$, and a substantial reduction in runtime, ranging from $58.6\%$ to $98.2\%$, compared to the reference methods.
Compared to the reference methods, LMF-based methods also exhibit a reduction in parameters, ranging from $45.1\%$ to $52.1\%$, which is attributed to the decoupling of computations from parameters, demonstrating the significance of our elegant design over merely stacking parameters.

\noindent \textbf{Qualitative Results.}
In Figure \ref{fig:qualitative_comparison}, we present a qualitative comparison for ASSR, using SwinIR as the encoder. The visual results demonstrate that our LMF-based methods achieve competitive visual quality across all scales compared to the reference methods. Remarkably, LMF-based methods achieve continuous and precise results while significantly reducing computational cost and runtime. This confirms that LMF effectively addresses the challenge of upsampling LR-HD features to HR-LD pixel values, eliminating the need for an HR-HD decoding function. More visual results are provided in the supplementary materials.
\section{Ablation Study}
In this section, we investigate the effect of each component in LMF. Both PSNR and MACs results on the DIV2K validation set are provided for a comprehensive evaluation, and all of the models use EDSR-b as the encoder. Notably, the CMSR algorithm is not used in Table \ref{tab:ablation_study_modulations} and Table \ref{tab:ablation_study_mlp}.

\noindent \textbf{Latent Modulation.}
In Table \ref{tab:ablation_study_modulations}, we evaluate the impact of latent modulation. Firstly, the comparison between LIIF and LIIF (tiny) reveals a significant decline in PSNR performance when shrinking the decoder from a 5-layer MLP with 256 dimensions to a 9-layer MLP with 16 dimensions. This performance drop is expected due to the substantial reduction in network parameters. However, our LM-LIIF model, which utilizes a 7-layer render MLP with 16 dimensions, outperforms LIIF in terms of PSNR, while consuming computational cost even lower than LIIF (tiny). Furthermore, we introduce a coarse-to-fine variant of LIIF, where both stages employ the same structure as LM-LIIF.
Unlike LM-LIIF, LIIF (C2F) fails to achieve competitive PSNR performance compared to LIIF, highlighting the critical role of latent modulation in reducing the dimensions of the render MLP. These results also validate that HR-HD computations are redundant for decoding LR-HD features.

In Table \ref{tab:ablation_study_modulations}, we also compare the performance of different types of latent modulations, namely the scale modulation, the shift modulation, and the combination of scale and shift modulations. The results demonstrate that LM-LIIF achieves the highest PSNR performance when both scale and shift modulations are employed, and the computational cost remains nearly identical.
Finally, we compare using the latent code (64 dimensions) from the encoder and the compressed latent code (16 dimensions) from the latent MLP as the input of the render MLP. The results indicate that using the compressed latent code can further reduce the computational cost for large scales. Based on these findings, we incorporate both scale and shift modulations as well as the compressed latent code in our final LMF models.

\begin{table}[t]
\centering
\footnotesize
\caption{Ablation study (PSNR(dB)/MACs on DIV2K validation set) on the latent MLP and the render MLP, using EDSR-b as the encoder, and LM-LIIF as the decoder.
}
\label{tab:ablation_study_mlp}
\begin{tabular}{l|l|cc|cc}
\hline
Latent & Render & \multicolumn{2}{c|}{In-scale} & \multicolumn{2}{c}{Out-of-scale} \bigstrut[t]\\
MLP & MLP & $\times2$ & $\times4$ & $\times8$ & $\times16$\bigstrut[t]\\
\hline

$2\times256$ & $2\times128$ & \makecell{34.66/\\236.91G} & \makecell{28.97/\\131.21G} & \makecell{25.37/\\104.57G} & \makecell{22.60/\\97.18G}\bigstrut[t]\\ 
\cline{3-6}
$2\times256$ & $3\times64$ & \makecell{34.67/\\232.18G} & \makecell{\textbf{29.00}/\\126.48G} & \makecell{\textbf{25.41}/\\99.85G} & \makecell{\textbf{22.63}/\\92.50G}\bigstrut[t]\\ 
\cline{3-6}
$2\times256$ & $5\times32$ & \makecell{\textbf{34.68}/\\197.37G} & \makecell{\textbf{29.00}/\\91.67G} & \makecell{\textbf{25.41}/\\65.11G} & \makecell{\textbf{22.63}/\\58.04G}\bigstrut[t]\\ 
\hline

$4\times256$ & $9\times16$ & \makecell{34.64/\\258.37G} & \makecell{28.98/\\87.79G} & \makecell{25.40/\\45.04G} & \makecell{22.62/\\34.13G}\bigstrut[t]\\ 
\cline{3-6}
$2\times256$ & $9\times16$ & \makecell{34.66/\\171.85G} & \makecell{28.99/\\66.16G} & \makecell{25.40/\\39.65G} & \makecell{22.62/\\32.79G}\bigstrut[t]\\ 
\cline{3-6}
\textbf{$2\times192$} & \textbf{$7\times16$} & \makecell{34.66/\\123.10G} & \makecell{\textbf{29.00}/\\49.91G} & \makecell{25.40/\\31.55G} & \makecell{\textbf{22.63}/\\26.77G} \bigstrut[t]\\
\cline{3-6}
$2\times128$ & $5\times16$ & \makecell{34.63/\\ \textbf{79.76G}} & \makecell{28.98/\\ \textbf{35.02G}} & \makecell{25.39/\\ \textbf{23.79G}} & \makecell{22.62/\\ \textbf{20.83G}}\bigstrut[t]\\ 
\hline
\end{tabular}
\vspace{-0.4cm}
\end{table}

\noindent \textbf{Network Structure.}
We evaluated different depths and dimensions of the latent MLP and the render MLP in Table \ref{tab:ablation_study_mlp}. Notably, the compressed latent code is not used to ensure better alignment of depths and dimensions between the two MLPs. For the latent MLP, we observe that decreasing the depth of the latent MLP from 4 to 2 yields nearly the same PSNR performance, which suggests that the latent MLP does not require a deep architecture or a large number of parameters.
However, when reducing the dimension of the latent MLP from 192 to 128, the PSNR performance begins to decrease. This observation confirms that HD computations are crucial for a powerful latent MLP.

In Table \ref{tab:ablation_study_mlp}, we investigate the depths and dimensions of the render MLP by fixing the dimensions of the latent MLP and the latent modulation at 256.
The results demonstrate that we can effortlessly reduce the dimension of the render MLP from 128 to 16, resulting in an impressive up to $66.26\%$ reduction in computational cost, without compromising performance. When the latent modulation possesses a fixed dimension, it is consistently viable to utilize a lower dimensional rendering process, thereby minimizing the computational burden.
All things considered, our final LM-LIIF model employs a 2-layer latent MLP with 208 dimensions (16 dimensions for the compressed latent code) and a 7-layer render MLP with 16 dimensions, optimizing computational efficiency without sacrificing performance.

\begin{table}[t]
\centering
\caption{Ablation study (PSNR(dB)/MACs of rendering/MACs of decoder on DIV2K validation set) on MSE thresholds of CMSR, using EDSR-b as the encoder, and LM-LIIF as the decoder.}
\label{tab:ablation_study_controllable_render}
\resizebox{\linewidth}{!}{
\begin{tabular}{l|cc|cc}
\hline
MSE & \multicolumn{2}{c|}{In-scale} & \multicolumn{2}{c}{Out-of-scale} \bigstrut[t]\\
threshold & $\times2$ & $\times4$ & $\times8$ & $\times16$\bigstrut[t]\\
\hline

$\times$ & \makecell{\textbf{34.65}/17.41G/\\125.32G} & \makecell{\textbf{28.99}/17.41G/\\44.38G} & \makecell{\textbf{25.40}/17.37G/\\24.10G} & \makecell{\textbf{22.62}/17.23G/\\18.90G} \bigstrut[t]\\
\cline{2-5}

$2\times10^{-5}$ & \makecell{\textbf{34.65}/16.30G/\\124.22G} & \makecell{\textbf{28.99}/13.67G/\\40.62G} & \makecell{\textbf{25.40}/12.41G/\\19.14G} & \makecell{\textbf{22.62}/3.48G/\\5.15G} \bigstrut[t]\\
\cline{2-5} 

$1\times10^{-4}$ & \makecell{34.62/13.63G/\\121.55G} & \makecell{28.98/11.71G/\\38.69G} & \makecell{25.39/8.35G/\\15.08G} & \makecell{\textbf{22.62}/2.50G/\\4.17G} \bigstrut[t]\\ 
\cline{2-5}

$5\times10^{-4}$ & \makecell{34.42/\textbf{10.35G}/\\ \textbf{118.27G}} & \makecell{28.93/\textbf{8.94G}/\\ \textbf{35.92G}} & \makecell{25.38/\textbf{6.08G}/\\ \textbf{12.81G}} & \makecell{22.61/\textbf{2.29G}/\\ \textbf{3.96G}} \bigstrut[t]\\ 
\hline
\end{tabular}
}
\vspace{-0.4cm}
\end{table}

\noindent \textbf{Controllable Multi-Scale Rendering.}
In Table \ref{tab:ablation_study_controllable_render}, we evaluate the MSE thresholds when creating the Scale2Mod table for CMSR.
Compared to the case without CMSR, a Scale2Mod table created with an MSE threshold of $2 \times 10^{-5}$ achieves the same PSNR performance while reducing computational cost by $20.58\%$ and $72.75\%$ for scale $\times 8$ and $\times 16$. With lower precision requirements, \ie, MSE thresholds of $1 \times 10^{-4}$ and $5 \times 10^{-4}$, CMSR effectively reduces the computational cost even further while maintaining maximum PSNR performance. This adaptability is advantageous for real-world applications as the precision can be adjusted during testing without the need for model retraining. However, CMSR exclusively reduces the computational cost of the render MLP, which limits the computational cost reduction for small scales.
Hence, in the Experiments section, we only employ CMSR with the MSE threshold $2 \times 10^{-5}$ for out-of-distribution scales. More details and results regarding CMSR are provided in the supplementary material.
\section{Conclusion}
In this paper, we propose a Latent Modulated Function (LMF), which realizes a computational optimal paradigm of continuous image representation. LMF utilizes an MLP in latent space to generate the latent modulation of each latent code. This enables a modulated MLP in render space to perform effective rendering at any resolution. Moreover, based on the correlation between modulation intensity and signal complexity, we propose a Controllable Multi-Scale Rendering (CMSR) algorithm, offering the flexibility to balance the rendering precision and efficiency during testing. Experiments demonstrate that converting INR-based ASSR methods to LMF leads to substantial reductions in computational cost, runtime, and parameters, while preserving performance. Better forms of latent modulations and more efficient decoder architectures will be explored in future work.
\section{Acknowledgement}
This work was supported by the National Natural Science Foundation of China under Grant No. 62071500; Supported by Shenzhen Science and Technology Program under Grant No. JCYJ20230807111107015.

{
    \small
    \bibliographystyle{ieeenat_fullname}
    \bibliography{main}
}

\end{document}